\documentclass{article}

\usepackage[preprint]{neurips_2025}

\usepackage[utf8]{inputenc} 
\usepackage[T1]{fontenc}    
\usepackage{hyperref}       
\usepackage{url}            
\usepackage{booktabs}       
\usepackage{amsfonts}       
\usepackage{nicefrac}       
\usepackage{microtype}      
\usepackage{xcolor}         

\usepackage{amsmath}
\usepackage{amssymb}
\usepackage{mathtools}
\usepackage{amsthm}
\theoremstyle{plain}
\usepackage{babel}
\usepackage[capitalize,noabbrev]{cleveref}

\theoremstyle{plain}
\newtheorem{theorem}{Theorem}[section]

\theoremstyle{definition}

\theoremstyle{remark}
\newtheorem{remark}[theorem]{Remark}
\usepackage{wrapfig,lipsum,booktabs}
\usepackage{subcaption}
\usepackage{microtype}
\usepackage{graphicx}
\usepackage{booktabs} 
\usepackage{bbm,multirow,tabularx}
\usepackage[ruled,vlined]{algorithm2e}

\usepackage{float}
\usepackage{arydshln}
\usepackage{amsmath}
\usepackage{amssymb}
\usepackage{mathtools}
\usepackage{amsthm}

\newcolumntype{Y}{>{\centering\arraybackslash}X}

\theoremstyle{plain}

\theoremstyle{plain}

\ifx\proof\undefined
\newenvironment{proof}[1][\protect\proofname]{\par
	\normalfont\topsep6\p@\@plus6\p@\relax
	\trivlist
	\itemindent\parindent
	\item[\hskip\labelsep\scshape #1]\ignorespaces
}{%
	\endtrivlist\@endpefalse
}
\providecommand{\proofname}{Proof}
\fi

\newcommand{\iid}{\overset{\mathrm{iid}}{\sim}}
\newcommand{\un}{\mathrm{un}}
\newcommand{\pos}{\mathrm{pos}}

\newcommand{\supp}{\mathrm{sup}}
\newcommand{\adv}{\mathrm{adv}}
\newcommand{\data}{\mathrm{data}}

\title{Generalization Bounds for Robust Contrastive Learning: From Theory to Practice}

\author{%
  Lam Tran\thanks{Equal contribution.} \\
  Qualcomm AI Research\thanks{Qualcomm AI Research is an initiative of Qualcomm Technologies, Inc.} \\
  \And
  Ngoc N. Tran$^{\ast}$ \\
  Vanderbilt University \\
  \And
  Hoang Phan$^\dagger$ \\
  New York University \\
  \And
  Anh Bui \\
  Monash University \\
  \AND
  Tung Pham \\
  Qualcomm AI Research \\
  \And
  Toan Tran \\
  Qualcomm AI Research \\
  \And
  Dinh Phung \\
  Monash University \\
  \And
  Trung Le \\
  Monash University \\
}

\begin{document}

\maketitle

\begin{abstract}
  Contrastive Learning first extracts features from unlabeled data, followed by linear probing with labeled data. Adversarial Contrastive Learning (ACL) integrates Adversarial Training into the first phase to enhance feature robustness against attacks in the probing phase. While ACL has shown strong empirical results, its theoretical understanding remains limited. Furthermore, while a fair amount of theoretical works analyze how the unsupervised loss can support the supervised loss in the probing phase, none has examined its role to the robust supervised loss. To fill this gap, our work develops rigorous theories to identify which components in the unsupervised training can help improve the robust supervised loss. Specifically, besides the adversarial contrastive loss, we reveal that the benign one, along with a global divergence between benign and adversarial examples can also improve robustness. Proper experiments are conducted to justify our findings; all code used in this work is available at \href{https://anonymous.4open.science/r/rosa}{this link}.
\end{abstract}
\section{Introduction}

Contrastive learning (CL) has become an essential technique in self-supervised learning (SSL) where positive and negative examples are created for each given anchor example \citep{he2019moco, misra2019contrastive, tian2019contrastive}.
Using CL, a feature extractor can learn to align representations of the anchors and their positive examples while contrasting which of their negative ones. A pioneering work in this area is SimCLR \citep{chen2020simple}, which proposed an efficient technique to train a feature extractor with CL. In SimCLR, positive examples are generated using random data augmentation techniques sampled from a pool $\mathcal{T}$, while negative examples are sampled from the data distribution. Subsequently, the InfoNCE loss~\citep{gutmann2010noise,oord2018representation} is employed to train a feature extractor by aligning the representations of positive pairs while contrasting those of negative examples.

The success of SimCLR and other CL techniques \citep{chuang2020debiased, robinson2020hard} has spurred interest in exploring the foundations of CL with InfoNCE loss. Prior works \citep{he2019moco, misra2019contrastive, tian2019contrastive} have served as the basis for many recent notable theoretical developments~\citep{psaunshi19a,wang2020understanding,wang2022chaos}.
However, none of these works has yet to investigate the robustness of CL.

The vulnerability of neural networks to imperceptibly small perturbations \citep{goodfellow2014explaining}
has posed a significant challenge in deploying them for safety-critical applications, such as autonomous driving \citep{caoccs19,8578273,zhao2019seeing}. Researchers have proposed various studies to enhance the robustness of trained networks against adversarial attacks \citep{madry2017towards, trades}, random noise \citep{zheng2016improving}, and corruptions \citep{hendrycks2019benchmarking}. In particular, the robustness of CL has been empirically investigated in previous works \citep{NEURIPS2020_robust1, NEURIPS2020_robust2, NEURIPS2020_robust3}.
Specifically, these works explored adversarial examples generated by modifying the InfoNCE loss to obtain robust unsupervised features for the second phase. While achieving promising empirical robust accuracy, these studies still fail to address why robustness transfers across CL training phases from a theoretical perspective.

In this work, we aim to provide a novel theoretical understanding of the robustness of CL. Our focus is on connecting the supervised adversarial loss in the second phase with the unsupervised loss in the first phase. We demonstrate that incorporating adversarial training in the first phase by creating adversarial examples that attack the InfoNCE loss can improve the robustness in the second phase and provide a theoretical justification for this phenomenon. Additionally, we show that utilizing a global view and enhancing the sharpness of the feature extractor \citep{foret2021sharpnessaware} can also strengthen the robustness in the second phase. Finally, we conduct empirical experiments to validate our theoretical findings and compare them with baselines.

To summarize, our contributions in this paper include:
\begin{itemize}
    \item We rigorously develop theoretical bounds on the adversarial loss in the second phase, identifying key factors that influence the robustness of the first phase. To the best of our knowledge, our work is \textbf{the first} to provide a theoretical treatment of this problem. In contrast, existing studies \cite{psaunshi19a, wang2020understanding, wang2022chaos, haochen2021provable} focus solely on the relationship between contrastive losses in the first phase and clean accuracy in the second phase.  
\item Our theoretical findings highlight several essential insights: \textbf{(i)} minimizing the InfoNCE loss on benign examples proves beneficial in enhancing both the robust and natural accuracies during the second phase, \textbf{(ii)} utilizing Sharpness-Aware Minimization (SAM) \citep{foret2021sharpnessaware} on the InfoNCE loss on benign examples offers further improvement in both robust and natural accuracies during the second phase, and \textbf{(iii)} employing adversarial attacks on the InfoNCE loss and pushing these adversarial examples away from their corresponding benign examples globally assists in enhancing robust accuracies during the second phase. It is noteworthy that (i) and (ii) contrast with the outcomes observed in standard adversarial training, where minimizing the benign loss can enhance natural accuracies but may compromise robust accuracies \citep{trades, rice2020overfitting}. We conduct experiments to validate our findings.
    \item Finally, leveraging our theoretical insights, we propose GSA-RSL (Globally Sharpness-Aware Robust Self-Supervised Learning). Experiments show it surpasses state-of-the-art baselines in both natural and robust accuracy, with significant gains in natural accuracy.
\end{itemize}
\section{Related work}
\subsection{Self-Supervised Learning}
\vspace{-1mm}

\textbf{Standard SSL} aims to extract meaningful representations from input data without relying on human annotations. Recent advances in SSL such as MoCo \citep{he2019moco}, SimCLR \citep{chen2020simple}, 
have demonstrated the effectiveness of these representations in various visual downstream tasks. The underlying principle of these methods is to learn representations that remain invariant to different data augmentations by maximizing the similarity of representations derived from different augmented versions of an image. 
Commonly used is CL with the InfoNCE loss to align positive pairs while spreading negative pairs apart \citep{he2019moco, misra2019contrastive, tian2019contrastive}. 

\textbf{Robust CL.} Machine learning models are highly vulnerable to adversarial perturbations \citep{szegedy2013intriguing, goodfellow2014explaining}, which, though imperceptible to humans, can cause critical errors in real-world systems. A promising approach to mitigate this issue is to develop a robust feature extractor \citep{NEURIPS2020_robust1, NEURIPS2020_robust2, NEURIPS2020_robust3, mao2019metric, fan2021does}, enabling strong classifiers even with simple linear layers.  \citep{NEURIPS2020_robust1, NEURIPS2020_robust2, NEURIPS2020_robust3} introduced robust SSL methods that train feature extractors with adversarial examples maximizing the InfoNCE loss, differing in their construction of positive/negative sets. \citep{NEURIPS2020_robust1} treated adversarial examples as positives, while \citep{NEURIPS2020_robust2} used two adversarial examples per anchor, doubling computation. In a supervised setting, \citep{bui2020improving, bui2021understanding} applied a similar regularization, leveraging class and adversarial/benign identities to align feature representations. \citep{xu2023enhancing} leveraged the technique of causal reasoning to interpret the ACL and propose adversarial invariant
regularization (AIR) to enforce independence from style factors. \citep{luo2023rethinking} proposed an augmentation schedule that gradually anneals from a strong augmentation to a weak one to benefit from both extreme cases for improving the robustness.

\textbf{Theory of SSL.} Driven by promising empirical results in CL, several studies have explored this learning paradigm from a theoretical perspective \citep{psaunshi19a, wang2020understanding, wang2022chaos}. More specifically, \citep{psaunshi19a} established a connection between supervised and unsupervised losses by using Rademacher complexity in binary classification. Moreover, \citep{wang2020understanding} examined the distribution of latent representations over the unit sphere and found that these representations tend to be uniformly distributed, encouraging alignment between positive examples and their anchors. Most recently, \citep{wang2022chaos} relaxed the conditional independence assumption in \citep{psaunshi19a} by introducing a milder condition, established a connection between supervised and unsupervised losses, and developed a new theory of augmentation overlap for CL. \citep{haochen2021provable} applied spectral decomposition—a classical approach for graph partitioning, also known as spectral clustering to the adjacency matrix defined on the population augmentation graph and studied theoretical properties of this contrastive representation.

\subsection{Flat Minima}

 Flat minimizers have been shown to enhance the generalization ability of neural networks by enabling them to discover wider local minima. This makes the model more resilient to differences between the training and testing datasets.
SAM \citep{foret2021sharpnessaware}, an effective flatness seeking method, has been successfully applied in various domains and tasks, such as meta-learning bi-level optimization \citep{abbas2022sharp}, federated learning \citep{qu2022generalized}, vision models \citep{chen2021vision}, language models \citep{bahri-etal-2022-sharpness} and domain generalization \citep{cha2021swad}.
It has also been used in supervised adversarial robustness \citep{wu2020adversarial}. In our work, we also leverage SAM, but extend the its theory to the unsupervised setting supporting
the InfoNCE loss.
\section{Problem Formulation and Notions}

In this section, we present the problem formulation of SSL and the notions used in our following theory development. We consider an $M$-class classification problem with the label set $\mathcal{Y}=\left\{ 1,2,...,M\right\} $. Given a class $c\in\mathcal{Y}$, the class-condition distribution for this class has the density function $p_{c}\left(x\right)=p\left(x\mid y=c\right)$ where $x\in\mathbb{R}^{d}$ specifies a data example. Therefore, the entire data distribution has the form
\begin{align}
p_{\text{data}}\left(x\right)=\sum_{c=1}^{M}\pi_{c}p\left(x\mid y=c\right)=\sum_{c=1}^{M}\pi_{c}p_{c}\left(x\right),
\end{align}
where $\pi_{c}=\mathbb{P}\left(y=c\right),c\in\mathcal{Y}$ is a class probability. The distribution of positive pairs $\left(x,x^{+}\right)$ over $\mathbb{R}^{d}\times\mathbb{R}^{d}$ is formulated as $p_{\text{pos}}\left(x,x^{+}\right)=\sum_{c=1}^{M}\pi_{c}p_{c}\left(x\right)p_{c}\left(x^{+}\right).    
$
It is worth noting that with the above equality, $p_{\text{pos}}\left(x,x^{+}\right)$ is relevant to the probability that $x,x^{+}\sim p_{\text{data}}$ possess the same label. Particularly, to form a positive pair $\left(x,x^{+}\right)$,
we first sample a class $c\sim\text{Cat}\left(\pi\right)$ from the categorical distribution with $\pi=\left[\pi_{c}\right]_{c=1}^{M}$, and then sample $x,x^{+}\sim p_{c}$. 

The general unsupervised InfoNCE~\citep{gutmann2010noise,oord2018representation} loss over the entire data and positive pair distributions is denoted as 
\begin{align}
\label{eq:L_un_general}
&\mathcal{L}_{\mathcal{D}_{\text{un}}}^{\text{un}}\left(\theta, p_{\text{pos}}\right)=\mathbb{E}_{\left(x,x^{+}\right)\sim p_{\text{pos}},x_{1:K}^{-} \overset{\text{iid}}{\sim}  p_{\text{data}}} 
 \Biggl[
  -\log\frac{\exp\big\{ \frac{f_{\theta}\left(x\right)\cdot f_{\theta}\left(x^{+}\right)}{\tau}\big\} }{\exp\big\{ \frac{f_{\theta}\left(x\right)\cdot f_{\theta}\left(x^{+}\right)}{\tau}\big\} +\frac{\beta}{K}\sum_{k=1}^{K}\exp\big\{ \frac{f_{\theta}\left(x\right)\cdot f_{\theta}\left(x_{k}^{-}\right)}{\tau}\big\} }\Biggr] ,
\end{align}
where $f_{\theta}$ with $\theta\in\Theta$ is a feature extractor, the operation $f_{\theta}(x) \cdot f_{\theta}(\widetilde{x})$ presents the inner product, $\tau>0$
is the temperature variable, $K$ is the number of used negative examples, and $\mathcal{D}_{\text{un}}$ denotes the distribution over $z=\left[x,x^{+},\left[x_{k}^{-}\right]_{k=1}^{K}\right]$
with $\left(x,x^{+}\right)\sim p_{\text{pos}},x_{1:K}^{-} \sim p_{\text{data}}$.
Note that $\beta \geq 0$ is a parameter and setting $\beta=K$ recovers
the original formula of CL.

Our ultimate goal is to minimize the \textit{general unsupervised InfoNCE loss}.
However, in reality, we work with a specific training set $\mathcal{S}=\Big\{ z_{i}=\Big[x_{i},x_{i}^{+},\big[x_{ik}^{-}\big]_{k=1}^{K}\Big]\Big\} _{i=1}^{N}$
where $z_{1:N}\sim\mathcal{D}_{\text{un}}$. The empirical unsupervised InfoNCE
loss over $\mathcal{S}$ is defined as
\begin{align}
\label{benign_loss}
&\mathcal{L}_{\mathcal{S}}^{\text{un}}\left(\theta, p_{\text{pos}}\right)=-\frac{1}{N}\times\sum_{i=1}^{N} \log H(x_i,x_i^+, [x_{ik}^-]_{k=1}^K),
\end{align}
where we have defined
\begin{align}
\label{eq:cl-i}
 H(x_i,x_i^+, [x_{ik}^-]_{k=1}^K)
 =\frac{\exp\big\{ \frac{f_{\theta}\left(x_{i}\right)\cdot f_{\theta}\left(x_{i}^{+}\right)}{\tau}\big\} }{\exp\big\{ \frac{f_{\theta}\left(x_{i}\right)\cdot f_{\theta}\left(x_{i}^{+}\right)}{\tau}\big\} +\frac{\beta}{K}\sum_{k=1}^{K}\exp\big\{ \frac{f_{\theta}\left(x_{i}\right)\cdot f_{\theta}\left(x_{ik}^{-}\right)}{\tau}\big\} }
\end{align}
SSL aims to minimize the \textit{empirical unsupervised InfoNCE loss}
over a specific training set $\mathcal{S}$ to learn an optimal feature extractor $f_{\theta^{*}}$, used in
the second phase for training a linear classifier on top of its extracted features. Given a feature extractor
$f_{\theta}$ and a linear classifier parameterized by a weight matrix $W$,
we define the general loss w.r.t. this couple as
\begin{align}
\mathcal{L}_{\mathcal{D}_{\text{sup}}}^{\text{sup}}\left(\theta,W\right)=\mathbb{E}_{\left(x,y\right)\sim\mathcal{D}_{\text{sup}}}\left[\tau_\text{CE}\left(Wf_{\theta}\left(x\right),y\right)\right],
\end{align}
where $\mathcal{D}_{\text{sup}}\left(x,y\right)=\pi_{y}p_{y}\left(x\right)$
is the data-label distribution and $\tau_\text{CE}\left(\cdot,\cdot\right)$
is the $\tau$-temperature cross-entropy loss (i.e., softmax with a temperature $\tau$ applied to logits before computing the cross-entropy loss). 

Given the fact that we aim to train the optimal linear classifier
in the second phase, we define the optimal general loss over all weight
matrices $W$ as   
\begin{align}
\mathcal{L}_{\mathcal{D}_{\text{sup}}}^{\text{sup}}\left(\theta\right)=\min_{W}\mathcal{L}_{\mathcal{D}_{\text{sup}}}^{\text{sup}}\left(\theta,W\right).
\end{align}
We now present the losses for the robust SSL. Given an attack $\mathbf{\bar a}$ (e.g., FGSM \citep{goodfellow2014explaining}, PGD \citep{madry2017towards}, TRADES \citep{trades}, or AutoAttack \citep{croce2020reliable}) and a weight matrix $W$, we denote $\mathbf{\bar a}_{W,\theta}(x) \in B_\epsilon(x) = \{x':\Vert x' - x \Vert \leq \epsilon \}$ as the \textit{adversarial example} obtained by using $\mathbf{\bar a}$ to attack the second phase classifier, formed by $f_\theta$ and $W$, on a benign example $x$ with a label $y$. Note that this should not be mistaken with $\mathbf{a}$, which will later denote the adversarial attack used in the first phase of CL. We define the adversarial loss using the adversary $\mathbf{\bar a}$ to attack $W$ and $f_\theta$ as follows:
\begin{equation}
\mathcal{L}_{\mathcal{D}_{\sup}}^{\adv}\left(W,\theta,\mathbf{\bar a}\right)=\mathbb{E}_{(x,y)\sim\mathcal{D}_{\supp}}\left[\tau_{\mathrm{CE}}\left(Wf_{\theta}\left(\mathbf{\bar a}_{W,\theta}\left(x\right)\right),y\right)^{\frac{1}{2}}\right]\label{eq:adv_loss_W_theta_a}
\end{equation}
Our next step is to minimize the above adversarial loss to define the best adversarial loss corresponding to the most robust linear classifier when using the adversary $\mathbf{\bar a}$ to attack the classifier in the second phase:
\begin{equation}
\mathcal{L}_{\mathcal{D}_{\sup}}^{\adv}\left(\theta,\mathbf{\bar a}\right)=\min_{W}\mathcal{L}_{\mathcal{D}_{\supp}}^{\adv}\left(W,\theta,\mathbf{\bar a}\right).\label{eq:adv_loss_theta_a}
\end{equation}

\section{Proposed Framework}
\subsection{Theory Development}
We first develop necessary theories to tackle the optimization problem of the adversarial loss $\mathcal{L}^{\adv}_{\mathcal{D}_{\supp}}(\theta, \mathbf{\bar a})$ whose all proofs can be found in Appendix. Let us denote $p_{\text{adv}}^{\mathbf{\bar a}_{W,\theta}} = \mathbf{\bar a}_{W, \theta} \# p_{\text{data}}$ as the adversarial distribution formed by using the attack transformation $\mathbf{\bar a}_{W,\theta}$ to transport the data distribution $p_{\text{data}}$. It is obvious by the definition that $p_{\text{adv}}^{\mathbf{\bar a}_{W,\theta}}$ consists of adversarial examples $\mathbf{\bar a}_{W,\theta}(x)$ with the benign example $x \sim p_{\text{data}}$. In addition, we define $q_{\text{data}} = f_\theta \# p_{\text{data}}$ and $q_{\text{adv}}^{\mathbf{\bar a}_{W,\theta}} = f_\theta \# p^{\mathbf{\bar a}_{W,\theta}}_{\text{adv}}$ as the benign and adversarial distributions on the latent space, respectively.

The following theorem indicates the upper bounds for the adversarial loss of interest $\mathcal{L}^{\adv}_{\mathcal{D}_{\supp}}(\theta, \mathbf{\bar a})$.

\begin{theorem}
    \label{thm:adv_loss_upper_main}
Consider the adversarial loss $\mathcal{L}^{\adv}_{\mathcal{D}_{\supp}}(\theta, \mathbf{\bar a})$.

i) We have the first upper bound on the data space
\begin{equation}
\begin{multlined}
\mathcal{L}_{\mathcal{D}_{\supp}}^{\adv}\left(\theta,\mathbf{\bar a}\right)\leq\min_{W}\Big\{ \mathcal{L}_{\mathcal{D}_{\supp}}^{\supp}\left(\theta,W\right)^{\frac{1}{2}} +\mathcal{L}_{\mathcal{D}_{\supp}}^{\supp}\left(\theta,W\right)^{\frac{1}{2}}D_{v}\left(p_{\adv}^{\mathbf{\bar a}_{W,\theta}},p_{\data}\right)^{\frac{1}{2}}\Big\} .\label{eq:1st_bound_data}
\end{multlined}
\end{equation}
ii) We have the second upper bound on the latent space
\begin{equation}
\begin{multlined}
\mathcal{L}_{\mathcal{D}_{\supp}}^{\adv}\left(\theta,\mathbf{\bar a}\right)\leq\min_{W}\Big\{ \mathcal{L}_{\mathcal{D}_{\supp}}^{\supp}\left(\theta,W\right)^{\frac{1}{2}} +\mathcal{L}_{\mathcal{D}_{\supp}}^{\supp}\left(\theta,W\right)^{\frac{1}{2}}D_{v}\left(q_{\adv}^{\mathbf{\bar a}_{W,\theta}},q_{\data}\right)^{\frac{1}{2}}\Big\} .\label{eq:2nd_bound_latent}
\end{multlined}
\end{equation}
Here we note that $D_v$ represents a $f$-divergence~\citep{ali1966general,nguyen2005divergences} with the corresponding convex function $v(t) = (t-1)^2$. 
\end{theorem}
Note that both bounds in Theorem \ref{thm:adv_loss_upper_main} have the same form but are defined on the data and latent spaces, respectively. Additionally, the upper bound in (\ref{eq:2nd_bound_latent}) reveals that to minimize the adversarial loss for the adversary $\mathbf{\bar a}$, we need to search the feature extractor $f_\theta$ and the weight matrix $W$ for minimizing the supervised loss $\mathcal{L}_{\mathcal{D}_{\supp}}^{\supp}\left(\theta,W\right)$ and the divergence $D_{v}\left(q_{\text{adv}}^{\mathbf{\bar a}_{W,\theta}},q_{\text{data}}\right)$ between the benign and adversarial distributions on the latent space.  

We now discover which losses should be involved in the first phase to minimize the upper bound in (\ref{eq:2nd_bound_latent}). To this end, given $x\sim p_{\text{data}}$ we denote \textit{the set of all possible attacks} as 
$\mathcal{A} =\left\{ \mathbf{a}\left(\cdot\mid x\right):\mathbf{a}\left(\cdot\mid x\right)\text{ has a support set over } B_{\epsilon}\left(x\right)\right\}$
and $q^{\mathbf{a}}=f_\theta \# (\mathbf{a}\#p_{\text{data}})\text{ for \ensuremath{\mathbf{a}\in\mathcal{A}}}$, where $B_\epsilon(x)$ specifies the ball with radius $\epsilon$ around $x$.  By its definition, the stochastic map ${\mathbf{\bar a}_{W, \theta}}$ can be represented by an element in $\mathcal{A}$ as long as $\mathbf{\bar a}_{W, \theta}(x) \in B_\epsilon(x),\forall x$. 
 
The following theorem assists us in converting the upper bound in (\ref{eq:2nd_bound_latent}) to the one directly in phase 1.
\begin{theorem}
    \label{thm:bound_phase_1}
    The adversarial loss $\mathcal{L}^{\adv}_{\mathcal{D}_{\supp}}(\theta, \mathbf{\bar a})$ can be further upper-bounded by
\begin{align}
\mathcal{L}_{\mathcal{D}_{\supp}}^{\adv}\left(\theta,\mathbf{\bar a}\right)\leq \left[\mathcal{L}_{\mathcal{D}_{\un}}^{\un}\left(\theta, p_{\pos}\right)+A\left(K,\beta\right)\right]^{\frac{1}{2}} \times  
\big[ 1 + \max_{\mathbf{a}\in\mathcal{A}} D_{v}\left(q^\mathbf{a},q_{\data}\right)^{\frac{1}{2}}\big]\label{eq:bound_phase_0}
\end{align}
where $A(K,\beta) = -O\big(\frac{1}{\sqrt{K}}\big)-\log\beta-O\big(\frac{1}{\beta}\big)$. 
\end{theorem}
\begin{remark} 
This theorem is both \textit{significant} and \textit{insightful}, as it establishes a clear connection between robust accuracy in the second phase and key factors in the first phase. Particularly, Inequality (\ref{eq:bound_phase_0}) points out that minimizing $\mathcal{L}_{\mathcal{D}_{\mathrm{un}}}^{\mathrm{un}}\left(\theta\right)$ and the divergence term $\max_{\mathbf{a} \in \mathcal{A}}D_{v}\left(q^{\mathbf{a}},q_{\text{data}}\right)$ is equivalent to minimizing an upper bound of
$\min_{\theta}\mathcal{L}_{\mathcal{D}_{\mathrm{sup}}}^{\mathrm{adv}}\left(\theta,\mathbf{\bar a}\right)$.
Moreover, this divergence term can be viewed as a \textit{global divergence} between the adversarial and benign distributions on the latent space. As a result, by pushing the adversarial examples in $q^\mathbf{a}$ globally close to the benign examples in $q_{\text{data}}$, we can strengthen the model's robustness to adversarial attacks in the second phase.    
\end{remark}
Unfortunately, minimizing $\mathcal{L}_{\mathcal{D}_{\mathrm{un}}}^{\mathrm{un}}\left(\theta, p_{\text{pos}}\right)$
directly is intractable due to the unknown general distribution $\mathcal{D}_{\mathrm{un}}$.
The following theorem resolves this issue and also signifies the concept
of sharpness for the feature extractor $f_{\theta}$.
\begin{theorem}
\label{thm:sharpness}Under mild conditions, with the probability at least $1-\delta$ over
the random choice of $\mathcal{S}\sim\mathcal{D}_{\mathrm{un}}^{N}$, we have
the following inequality 
\begin{align}
&\mathcal{L}_{\mathcal{D}_{\mathrm{un}}}^{\mathrm{un}}\left(\theta, p_{\pos}\right)  \leq\max_{\theta':\Vert\theta'-\theta\Vert<\rho}\mathcal{L}_{\mathcal{S}}^{\mathrm{un}}\left(\theta'\right)+ \frac{1}{\sqrt{N}}\Big[\log\frac{1}{\delta}+ 
\frac{T}{2}\log\Big(1+\frac{\Vert{\theta} \Vert^{2}}{T\sigma^{2}}\Big) 
+\frac{L^{2}}{8}+2L + O\big(\log(N+T)\big)\Big]
\end{align}
where $L=\frac{2}{\tau}+\log(1+\beta)$, $T$ is the number of parameters
in $\theta$, and $\sigma=\rho[\sqrt{T}+\sqrt{\log\left(N\right)}]^{-1}$.
\end{theorem}

The proof in \citep{foret2021sharpnessaware} relies on the McAllester PAC-Bayesian bound \citep{mcallester1999pac}, limiting it to 0-1 loss in binary classification. In contrast, we introduce the first sharpness-aware theory for SSL feature extractors, using the PAC-Bayesian bound \citep{JMLR:v17:15-290} to handle the more general InfoNCE loss. Leveraging Theorems \ref{thm:bound_phase_1} and \ref{thm:sharpness}, we derive the following result.  

\begin{theorem}
\label{thm:sharpness_div_bound}
Under mild conditions, with the probability at least $1-\delta$ over
the random choice of $\mathcal{S}\sim\mathcal{D}_{\mathrm{un}}^{N}$, we have the following inequality
\begin{align}
& \mathcal{L}_{\mathcal{D}_{\supp}}^{\adv}\left(\theta,\mathbf{\bar a}\right)\leq\Big[
\max_{\Vert\theta'-\theta\Vert\leq\rho}\mathcal{L}_{S}^{\un}\left(\theta', p_{\pos}\right)+B(T,N,\delta,\Vert\theta\Vert) 
 +A\left(K,\beta\right)\Big]^{\frac{1}{2}}\Big[1 + \max_{\mathbf{a}\in\mathcal{A}}D_{v}\left(q^\mathbf{a},q_{\data}\right)^{\frac{1}{2}}\Big],\label{eq:sharpness_div_bound}
\end{align}
where we have defined
\begin{align*}
B(T,N,\delta,\Vert\theta\Vert)=\frac{1}{\sqrt{N}}\Big[\frac{T}{2}\log\Big(1+\frac{\Vert\theta\Vert^{2}}{T\sigma^{2}}\Big)+\log\frac{1}{\delta}+\frac{L^{2}}{8} 
+ 2L+O\left(\log\left(N+T\right)\right)\Big].
\end{align*}
\vspace{-4mm}
\end{theorem}
\begin{remark}    Inequality (\ref{eq:sharpness_div_bound}) 
shows that minimizing the adversarial loss $\mathcal{L}_{\mathcal{D}_{\supp}}^{\adv}\left(\theta,\bar{\mathbf{a}}\right)$ with respect to $\theta$ requires minimizing the \textit{sharpness-aware unsupervised InfoNCE loss} $\max_{\Vert\theta'-\theta\Vert\leq\rho}\mathcal{L}_{S}^{\un}\left(\theta'\right)$ and the divergence
 $\max_{\mathbf{a} \in \mathcal{A}}D_{v}\left(q^\mathbf{a},q_{\data}\right)$. While
 the former term
 is straightforward, the divergence term $\max_{\mathbf{a} \in \mathcal{A}}D_{v}\left(q^\mathbf{a},q_{\text{data}}\right)$ requires further derivations, which will be clarified in the next section. 
\end{remark}
We now develop another upper bound for the adversarial loss $\mathcal{L}_{\mathcal{D}{\supp}}^{\adv}\left(\theta,\mathbf{\bar a}\right)$. Given  a stochastic map $\mathbf{a} \in \mathcal{A}$, we define the hybrid positive $p_{\pos}^{\mathbf{a}}(x,x^{+})$ distribution induced by $\mathbf{a}$ as
 $
p_{\text{pos}}^{\mathbf{a}}(x,x^{+})=\sum_{c=1}^{M}\pi_{c}p_{c}\left(x\right)p_{c}^{\mathbf{a}}\left(x^{+}\right),$ where $p^\mathbf{a}_c = \mathbf{a}\#p_c$. Additionally, by its definition, the hybrid positive distribution $p_{\text{pos}}^{\mathbf{a}}(x,x^{+})$ consists of the positive pairs of a benign example $x$ with the label $c$ and an adversarial example $x^+$ formed by applying the stochastic map $\mathbf{a}$ to another benign example $x'$. The following theorem gives us another upper bound on the adversarial loss $\mathcal{L}_{\mathcal{D}{\supp}}^{\adv}\left(\theta,\mathbf{\mathbf{\bar a}}\right)$. 
\begin{theorem}
\label{thm:bound_phase_1_direct}
The adversarial loss $\mathcal{L}^{\adv}_{\mathcal{D}_{\supp}}(\theta, \mathbf{\bar a})$ can be further upper-bounded by
\begin{align}\label{eq:bound_phase_1}
&\mathcal{L}_{\mathcal{D}_{\supp}}^{\adv}\left(\theta,\mathbf{\bar a}\right)\leq \Big[ \max_{\mathbf{a} \in \mathcal{A}} \Big\{\mathcal{L}_{\mathcal{D}_{\text{un}}}^{\un}\left(\theta, p^\mathbf{a}_{\pos}\right)  + \Big(\frac{1}{\tau} + \exp\big\{ \frac{1}{\tau}\big\}\Big) D_{u}\left(q^{\mathbf{a}},q_{\data}\right)  \Big\}+A\left(K,\beta\right)\Big]^{\frac{1}{2}},
\end{align}
where $A(K,\beta) = -O\big(\frac{1}{\sqrt{K}}\big)-\log\beta-O\big(\frac{1}{\beta}\big)$, $D_u$ is a $f$-divergence with the convex function $u(t) = |t-1|$, and $\mathcal{L}_{\mathcal{D}_{\un}}^{\un}\left(\theta, p^\mathbf{a}_{\pos}\right)$ is similar to $\mathcal{L}_{\mathcal{D}_{\un}}^{\un}\left(\theta, p_{\pos}\right)$ in Eq. (\ref{eq:L_un_general}) except that the positive pairs $(x, x^+) \sim p^\mathbf{a}_{\pos}$.  
\end{theorem}
\begin{remark}
Theorem \ref{thm:bound_phase_1_direct} indicates that we need to find the adversary $\mathbf{a}$ that maximizes the InfoNCE loss over the hybrid adversarial positive distribution $p^\mathbf{a}_{\text{pos}}$ and the divergence on the latent space $D_u(q^\mathbf{a}, q_{\text{data}})$. This allows us to create strong adversarial examples that cause high InfoNCE values (i.e., $z^\mathbf{a} = f_\theta(\mathbf{a}(x))$ is locally far from $z = f_\theta(x)$), while the distribution of adversarial examples is globally far from that of benign examples on the latent space. Eventually, the feature extractor $f_\theta$ is updated to minimize both the local distances and the global divergence.       
\end{remark}
Additionally, the adversarial InfoNCE term $\mathcal{L}_{\mathcal{D}_{\un}}^{\un}\left(\theta, p^\mathbf{a}_{\pos}\right)$ has the form of
\begin{align}
\mathcal{L}_{\mathcal{D}_{\un}}^{\un}\left(\theta,p_{\text{pos}}^{\mathbf{a}}\right) 
&=\mathbb{E}_{\left(x,x^{+}\right)\sim p_{\text{pos}}^{\mathbf{a}},x_{1:K}^{-}\iid p_{\text{data}}}[
-\log H(x,x^+,x_{1:K}^-)] \nonumber\\
& = \mathbb{E}_{\left(x,x^{+}\right)\sim p_{\text{pos}},x_{1:K}^{-}\iid p_{\text{data}}} [ 
-\log H(x,\mathbf{a}(x^+),x_{1:K}^-)],
\label{eq:adv_InfoNCE}
\end{align}
which means that given the positive pair $(x,x^+)$ having the same label, we find the adversarial example $\mathbf{a}(x^+)$ for $x^+$ that maximizes the InfoNCE loss as in Eq. (\ref{eq:adv_InfoNCE}). Note that according to Eq. (\ref{eq:bound_phase_1}), we also need to seek adversarial examples $\mathbf{a}(x^+)$ to globally maximize a $f$-divergence between the benign and adversarial distributions.  

Using the upper bounds in Theorems \ref{thm:sharpness_div_bound} and \ref{thm:bound_phase_1_direct}, we arrive at the final upper bound stated in the following theorem.
\begin{theorem}
    \label{thm:final_bound}
    For any $0\leq \lambda \leq 1$, the adversarial loss $\mathcal{L}^{\adv}_{\mathcal{D}_{\supp}}(\theta, \mathbf{\bar a})$ can be further upper-bounded by
\begin{alignat}{2}
\vspace{5mm}
     \mathcal{L}_{\mathcal{D}_{\supp}}^{\adv}\left(\theta,\mathbf{\bar a}\right) &\leq  
    \lambda\Big[\max_{\mathbf{a} \in \mathcal{A}} \Big\{\mathcal{L}_{\mathcal{D}_{\text{un}}}^{\un}\left(\theta, p^\mathbf{a}_{\pos}\right) + \Big(\frac{1}{\tau} + \exp\big\{ \frac{1}{\tau}\big\}\Big) D_{u}\left(q^{\mathbf{a}},q_{\data}\right)  \Big\}  
    +A\left(K,\beta\right)\Big]^{\frac{1}{2}}
    \nonumber \\ &+(1-\lambda)\Big[
    \max_{\Vert\theta'-\theta\Vert\leq\rho}\mathcal{L}_{S}^{\un}\left(\theta', p_{\pos}\right)+A\left(K,\beta\right) 
    +B(T,N,\delta,\Vert\theta\Vert)\Big]^{\frac{1}{2}}\Big[1 + \max_{\mathbf{a}\in\mathcal{A}}D_{v}\left(q^\mathbf{a},q_{\data}\right)^{\frac{1}{2}}\Big]. \label{eq:final_bound_phase} 
\end{alignat}
\end{theorem}

\begin{remark}
\label{full_remark}
    To summarize our theoretical findings:
    \begin{itemize}
        \item Minimizing the InfoNCE loss on the benign examples  $\mathcal{L}_{S}^{\un}\left(\theta, p_{\pos}\right)$ assists us in lowering down the adversarial loss in the second phase. Moreover, the SAM for the InfoNCE loss on the benign examples $\max_{\Vert\theta'-\theta\Vert\leq\rho}\mathcal{L}_{S}^{\un}\left(\theta', p_{\pos}\right)$ also supports us in reducing the adversarial loss in the second phase.
        \item \sloppy Minimizing the InfoNCE loss on the adversarial examples formed by attacking the benign examples on the InfoNCE loss $\max_{\mathbf{a} \in \mathcal{A}} \mathcal{L}_{\mathcal{D}_{\text{un}}}^{\un}\left(\theta, p^{\mathbf a}_{\pos}\right)$ helps strengthen the adversarial robustness in the second phase, as commonly done in prior works. Moreover, our theory highlights from a game theory perspective that minimizing the divergence between the benign examples and adversarial ones found by maximizing such divergence $D_{u}\left(q^{\mathbf{a}},q_{\data}\right)$ also enhances robustness.
        To this end, the adversarial examples are found by maximizing the InfoNCE loss and global divergence to their corresponding benign examples. 
        In our implementation, we relax $D_u$ to the Jensen-Shannon divergence $D_{JS}$ to utilize generative adversarial network (GAN) \citep{goodfellow2020generative} for quantifying the global divergence. 
 
    \end{itemize}
      
\end{remark}

\subsection{Practical Method}
This section explains how to harvest the developed theories to conduct a practical method for robust SSL. We emphasize that our main objective is not to seek state-of-the-art methods, but to demonstrate that the finding terms from the theories assist us in improving the robustness of SSL.   

Given a mini-batch $B = \{x_1,\dots, x_b\}$ of the anchor examples, we apply random transformations from a pool of transformation $\mathcal{T}$ to create the corresponding positive pairs $(\Tilde{x}_1, \Tilde{x}_1^+)$, $(\Tilde{x}_2, \Tilde{x}_2^+)$,..., $(\Tilde{x}_{b}, \Tilde{x}_{b}^+)$ where $\Tilde{x}_i = t(x_i)$ and $\Tilde{x}_{i}^+= t'(x_i)$ with $t,t' \sim \mathcal{T}$. Hinted by Eq. (\ref{eq:adv_InfoNCE}), we formulate the adversarial InfoNCE loss term 
$\mathcal{L}_{\mathcal{D}_{\text{un}}}^{\un}\left(\theta, p^\mathbf{a}_{\pos}\right)=\max_{\forall i:\tilde{x}_{i}^{+,\mathbf{a}}\in B_{\epsilon}\left(\tilde{x}_{i}^{+}\right)}\{ -\frac{1}{b} \times \sum_{i=1}^{b}\log H(\tilde{x}_i,\tilde{x}_i^{+,\mathbf{a}},\tilde{x}_k^-)\}$,
where we denote $\tilde{x}_{i}^{+,\mathbf{a}}$ as the adversarial example of $\tilde{x}_{i}^{+}$ within its $\epsilon$-ball.

We relax the global term $D_{u}\left(q^{\mathbf{a}},q_{\data}\right)$ to $D_{JS}\left(q^{\mathbf{a}},q_{\data}\right)$ so GAN can be applied to realize it. In specific, from \citep{goodfellow2020generative}, we have the inequality
\begin{align}
    &D_{JS}\left(q^{\mathbf{a}},q_{\data}\right) \geq  
    \frac{1}{2}\max_{D}\bigg\{\mathbb{E}_{\tilde{x}^+\sim q_{data}}[\log D(f_{\theta}(\tilde{x}^+))] 
    +\mathbb{E}_{\tilde{x}^{+,\boldsymbol{a}}\sim q^{\boldsymbol{a}}}[\log (1-D(f_{\theta}(\tilde{x}^{+,\boldsymbol{a}})))]+2\log2\bigg\},
    \label{eq:JS_ineq}
\end{align}
where $D$ is the discriminator that distinguishes samples from $q_{data}$ and $a^{\boldsymbol{a}}$. Furthermore, this inequality is tight if the  search space for $D$ has members to approach $D^*(z)=\frac{q_{data}(z)}{q_{data}(z)+q^{\boldsymbol{a}}(z)}$ up to any precision level or contains $D^*$. Therefore, by assuming a strong family function of $D$ and the convergence of training $D$, $D_{JS}\left(q^{\mathbf{a}},q_{\data}\right)$ can be approximated by the R.H.S of Eq. \ref{eq:JS_ineq}.
Now, we can combine the local and global views to generate adversarial examples
\begin{align}
\max_{\forall i:\tilde{x}_{i}^{+,\mathbf{a}}\in B_{\epsilon}\left(\tilde{x}_{i}^{+}\right)} \bigg\{ -\frac{1}{b}\sum_{i=1}^{b}\log H(\tilde{x}_i,\tilde{x}_i^{+,\mathbf{a}},\tilde{x}_{ik}^-) + 
\lambda_{\text{global}} D_\text{loss}
\bigg\},
\label{eq:generate_xa}
\end{align}
where $\lambda_{\text{global}} \geq 0$ is a parameter to trade-off between the local and global views, and $D_{loss}$ is defined as 
\begin{align}
    D_\text{loss}= \sum_{i=1}^{b}\log D\left(f_{\theta}\left(\tilde{x}_{i}^{+}\right)\right) +\sum_{i=1}^{b}\log (1- D\left(f_{\theta}\left(\tilde{x}_{i}^{+,\mathbf{a}}\right)\right)).
    \label{eq:GAN}
\end{align}
Note again that in Eq. (\ref{eq:generate_xa}), we indeed push the batch adversarial examples $\tilde{x}_{i}^{+,\mathbf{a}}$ globally far away from the batch of benign examples $\tilde{x}_{i}$ by maximizing $D_{loss}$.
Then, the discriminator $D$ is updated to distinguish the batch adversarial examples $\tilde{x}_{i}^{+,\mathbf{a}}$ and the batch of benign examples $\tilde{x}_{i}$ by maximizing the $D_\text{loss}$. Finally, given the batch of adversarial examples, we update the feature extractor $f_\theta$ as
\begin{equation}
\begin{multlined}
\min_{\theta}\bigg\{\frac{-1}{b}  \sum_{i=1}^{b}\log H(\tilde{x}_i,\tilde{x}_i^{+,\mathbf{a}},\tilde{x}_{ik}^-)
-\lambda_{\text{benign}}\mathcal{L}^{\text{un}}_{B}(\theta, p_{\text{pos}}) +
\lambda_{\text{global}}D_\text{loss}
\bigg \}.\label{eq:benign_model_update}
\end{multlined}
\end{equation}
Specifically, in addition to the adversarial InfoNCE, we incorporate the benign InfoNCE (cf. Eq. (\ref{benign_loss})) on the batch $B$ of benign examples with the trade-off parameter $\lambda_{\text{benign}}$. We simultaneously minimize $D_{loss}$ term w.r.t. $\theta$ to move the batch of adversarial examples globally closer to the batch of benign examples.

Note that we apply SAM \citep{foret2021sharpnessaware} to the InfoNCE loss on the benign examples. The only difference in computation is that we find the perturbed model $\theta^\mathbf{a}$ (i.e., $\theta^\mathbf{a}= \theta+ \rho\frac{\nabla_{\theta}\mathcal{L}_{B}^{\text{un}}\left(\theta,p_{\text{pos}}\right)}{\Vert\nabla_{\theta}\mathcal{L}_{B}^{\text{un}}\left(\theta,p_{\text{pos}}\right)\Vert}$) that maximizes the InfoNCE loss $\mathcal{L}^{\text{un}}_{B}(\theta, p_{\text{pos}})$ on the benign examples, then use the gradient at $\theta^\mathbf{a}$ for this term when updating $\theta$ during optimization.
It is worth noting that the first and third terms in Eq. (\ref{eq:benign_model_update}) are optimized using their gradients at $\theta$. The entire training process for our method is presented in Algorithm~\ref{algo:global-local}, and implementation details can be found in Appendix.


\begin{algorithm}[!h]
\caption{Algorithm for our GSA-RSL}
\label{algo:global-local}
\SetAlgoLined
\SetKwInOut{Input}{Input}
\Input{Training set $\mathcal{S}$, pool of augmentations $\mathcal{T}$, main network $f_{\theta}$, discriminator $D_{\phi}$, $\epsilon$-ball $B_{\epsilon}(.)$, 
number of epochs $E$.}

\SetKwInOut{Result}{Output}
\Result{$f_\theta$}


\For {$e$ in range($E$)} 
{
\ForEach {mini-batch $\mathcal{B} \subset \mathcal{S}$ }
{
1. Sample two augmentations $t, t' \sim \mathcal{T} $ and \\
get two augmentation views $\tilde{x}_i = t(x_i), \quad  
\tilde{x}_i^+ = t'(x_i) \quad \forall x_i \in \mathcal{B}$. \\
2. Generate adversarial images $\{\tilde{x}_i^{+,\mathbf{a}}\}_{i=1}^\mathcal{B}$ with Eq. (\ref{eq:generate_xa}).
\\
3. Update discriminator's parameters $\phi$ by maximizing the log-likelihood in Eq. (\ref{eq:GAN}).
\\
4. Update main model $\theta$ with Eq. (\ref{eq:benign_model_update}) using SAM by finding the perturbed model $\theta^\mathbf{a}$ for the InfoNCE loss and using the model at $\theta^\mathbf{a}$ to update the main model $\theta$ for this term, whereas the first and third terms still use the gradients at $\theta$.
}
}
\end{algorithm}
\vspace{-4mm}


.

\section{Experiments}

\subsection{Experimental Settings}

\textbf{General Settings:}
In this section, we verify the effectiveness of our method by conducting experiments on CIFAR-10, CIFAR-100 \citep{krizhevsky2009learning}, and STL-10 \citep{coates2011stl10} datasets, whose
details for the pre-process and augmentations are provided in the appendix. 
We use ResNet-18 \citep{he2016deep} for our feature extractor backbone and a simple single-layer net for the discriminator. 

\textbf{SSL Settings:}
For the contrastive learning process, we follow the settings generally used in existing works \citep{robinson2020hard,chuang2020debiased}, which can be found in the appendix.
In the linear evaluation (LE) phase, we drop the projection head and instead attach a linear classifier to our feature extractor and train it in a supervised manner. We also adversarially train this classifier, denoted as AT-LE. At evaluation, the discriminator is not used in the adversarial generation process.


\begin{table*}[!ht]
\caption{\small Clean and robust accuracies under $\ell_\infty$, $\ell_2$-PGD and AutoAttack, 
under the two settings of linear evaluation.}
\label{main-table}
\centering
\begin{tabularx}{\textwidth}{cc*{8}Y}
\toprule
&\multirow{3}{*}{Method} & \multicolumn{4}{c}{CIFAR-10} & \multicolumn{4}{c}{CIFAR-100} \\
\cmidrule(lr){3-6} \cmidrule(lr){7-10} 
 && Clean & $\ell_\infty$ & $\ell_2$ & AA & Clean & $\ell_\infty$ & $\ell_2$ & AA    \\
\midrule      			
\multirow{5}{*}{LE} 
&RoCL & $76.35$ & $38.52$ & $52.10$ & $34.23$ & $37.30$ & $18.14$ & $24.40$ & $13.69$ \\
&CLAE & $76.53$ & $38.77$ & $53.32$ & $34.62$ & $37.92$ & $17.76$ & $24.37$ & $13.31$ \\
&\textbf{Ours} & $\textbf{85.08}$ & $\textbf{41.69}$ & $\textbf{56.34}$ & $\textbf{38.60}$ & $\textbf{54.21}$ & $\textbf{18.58}$ & $\textbf{27.05}$ & $\textbf{14.37}$ \\
\cmidrule(lr){2-10} 
&AdvCL & $86.53$ & $43.55$ & $57.70$ & $39.76$ & $49.62$ & $20.67$ & $27.91$ & $16.54$ \\
&\textbf{AdvCL+Ours} & $\textbf{90.03}$ & $\textbf{44.95}$ & $\textbf{59.63}$ & $\textbf{40.82}$ & $\textbf{57.09}$ & $\textbf{21.15}$ & $\textbf{29.88}$ & $\textbf{16.81}$ \\
\midrule
\multirow{5}{*}{AT-LE} 
&RoCL & $71.11$ & $43.39$ & $54.14$ & $37.07$ & $32.29$ & $19.15$ & $23.86$ & $14.07$ \\
&CLAE & $72.54$ & $43.45$ & $55.28$ & $37.70$ & $32.69$ & $18.96$ & $23.89$ & $14.01$ \\
&\textbf{Ours} & $\textbf{81.41}$ & $\textbf{46.19}$ & $\textbf{58.31}$ & $\textbf{41.02}$ & $\textbf{42.72}$ & $\textbf{20.09}$ & $\textbf{26.20}$ & $\textbf{15.35}$ \\
\cmidrule(lr){2-10} 
&AdvCL & $83.26$ & $47.59$ & $59.16$ & $42.06$ & $42.48$ & $22.21$ & $28.48$ & $17.26$ \\
&\textbf{AdvCL+Ours} & $\textbf{87.81}$ & $\textbf{48.46}$ & $\textbf{61.18}$ & $\textbf{42.85}$ & $\textbf{48.87}$ & $\textbf{23.61}$ & $\textbf{30.55}$ & $\textbf{17.90}$ \\
\bottomrule

\end{tabularx}
\end{table*}

\begin{figure*}[!ht]
    \centering    
    \begin{subfigure}[b]{0.5\textwidth}
        \centering
        {\includegraphics[width=0.99\linewidth, scale=0.9]{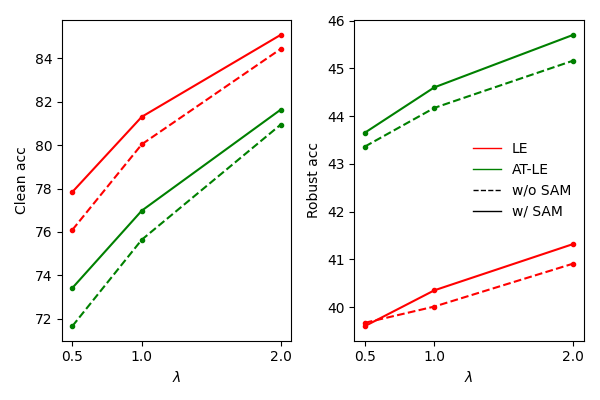}}%
        \vspace{-3mm}
        \caption{Impact of SAM.}
        \label{fig:SAM}
    \end{subfigure}%
    \hfill
    \begin{subfigure}[b]{0.5\textwidth}
        \centering
        {\includegraphics[width=0.99\linewidth, scale=0.9]{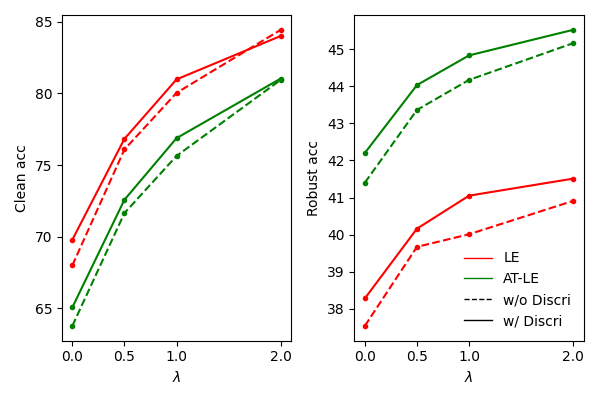}}%
        \vspace{-3mm}
        \caption{Impact of Discriminator.}
        \label{fig:D}
    \end{subfigure}%
    \caption{\small Clean \& robust accuracy with/out our method's components, across different weighting for the benign loss term.}
\label{fig:ablation}
\end{figure*}

\begin{table}[t]
    \centering
    
\begin{tabularx}{\columnwidth}{c*{5}Y}
\toprule
& \multirow{2}{*}{Method} & \multicolumn{4}{c}{CIFAR-10}  \\
\cmidrule(lr){3-6} 
 && Clean & $\ell_\infty$ & $\ell_2$ & AA     \\
\midrule      			
\multirow{8}{*}{LE} 
&ACL-DS & $76.63$ & $36.61$ & $52.00$ & $34.40$  \\
&\textbf{+Ours} & $\textbf{77.34}$ & $\textbf{37.51}$ & $\textbf{52.32}$ & $\textbf{35.58}$  \\
\cmidrule(lr){2-6} 
&ACL-Dyn & $78.31$ & $44.65$ & $57.27$ & $42.13$ \\
&\textbf{+Ours} & $\textbf{78.43}$ & $\textbf{45.93}$ & $\textbf{58.53}$ & $\textbf{43.30}$ \\
\cmidrule(lr){2-6} 
&ACL-AIR & $77.05$ & $36.18$ & $51.39$ & $33.90$ \\
&\textbf{+Ours} & $\textbf{78.00}$ & $\textbf{37.38}$ & $\textbf{52.06}$ & $\textbf{35.31}$ \\
\cmidrule(lr){2-6} 
&Dyn-AIR & $78.05$ & $44.80$ & $57.67$ & $42.43$ \\
&\textbf{+Ours} & $\textbf{78.59}$ & $\textbf{45.62}$ & $\textbf{58.07}$ & $\textbf{42.96}$ \\
\midrule      			
\multirow{8}{*}{AT-LE} 
&ACL-DS & $72.94$ & $42.24$ & $54.61$ & $38.07$  \\
&\textbf{+Ours} & $\textbf{73.85}$ & $\textbf{43.16}$ & $\textbf{54.92}$ & $\textbf{38.94}$  \\
\cmidrule(lr){2-6} 
&ACL-Dyn & $76.07$ & $48.34$ & $59.05$ & $44.49$ \\
&\textbf{+Ours} & $\textbf{76.37}$ & $\textbf{48.76}$ & $\textbf{59.08}$ & $\textbf{45.18}$ \\
\cmidrule(lr){2-6} 
&ACL-AIR & $73.38$ & $42.21$ & $54.19$ & $38.19$ \\
&\textbf{+Ours} & $\textbf{74.25}$ & $\textbf{42.94}$ & $\textbf{54.63}$ & $\textbf{39.00}$ \\
\cmidrule(lr){2-6} 
&Dyn-AIR & $76.65$ & $48.44$ & $59.04$ & $44.66$ \\
&\textbf{+Ours} & $\textbf{76.69}$ & $\textbf{48.83}$ & $\textbf{59.36}$ & $\textbf{45.21}$ \\
\bottomrule\\
\end{tabularx}
\caption{Compatibility with recent baselines.}
\label{new-baseliens-table}
\end{table}

\begin{minipage}{0.5\textwidth}
\small
\centering
\setlength\tabcolsep{0pt}
\begin{tabularx}{\columnwidth}{cc*{4}Y}
\toprule
& \multirow{2}{*}{Method} & \multicolumn{4}{c}{CIFAR-10 $\rightarrow$ STL-10}  \\
\cmidrule(lr){3-6} 
 && Clean & $\ell_\infty$ & $\ell_2$ & AA     \\
\midrule      			
\multirow{5}{*}{LE} 
&RoCL & $57.74$ & $28.04$ & $39.54$ & $23.44$  \\
&CLAE & $57.64$ & $26.55$ & $38.60$ & $22.50$  \\
&\textbf{Ours} & $\textbf{69.83}$ & $\textbf{28.95}$ & $\textbf{41.58}$ & $\textbf{24.61}$ \\
\cmidrule(lr){2-6} 
&AdvCL & $69.01$ & $28.67$ & $39.99$ & $24.06$ \\
&\textbf{AdvCL+Ours} & $\textbf{74.19}$ & $\textbf{29.35}$ & $\textbf{42.42}$ & $\textbf{24.82}$ \\
\midrule
\multirow{5}{*}{AT-LE} 
&RoCL & $54.08$ & $31.71$ & $41.11$ & $25.74$  \\
&CLAE & $53.91$ & $30.54$ & $40.16$ & $24.95$  \\
&\textbf{Ours} & $\textbf{64.64}$ & $\textbf{34.01}$ & $\textbf{43.66}$ & $\textbf{28.24}$ \\
\cmidrule(lr){2-6} 
&AdvCL & $62.85$ & $34.46$ & $43.66$ & $27.49$ \\
&\textbf{AdvCL+Ours} & $\textbf{70.66}$ & $\textbf{35.53}$ & $\textbf{46.55}$ & $\textbf{28.07}$ \\
\bottomrule
\end{tabularx}
\captionof{table}{Transferability from CIFAR-10 to STL-10.}
\label{stl-table}
\end{minipage}~
\begin{minipage}{0.48\textwidth}
\small
\centering
\centering
\begin{tabularx}{\linewidth}{c*{4}Y}
\toprule
\multirow{2}{*}{Method} & \multicolumn{2}{c}{LE} & \multicolumn{2}{c}{AT-LE} \\
\cmidrule(lr){2-3} \cmidrule(lr){4-5}
& Clean & $\ell_\infty$ & Clean & $\ell_\infty$    \\
\midrule
SimCLR & $84.44$ & $40.91$ & $80.95$ & $45.16$ \\
Only SAM & $84.74$ & $41.04$ & $81.35$ & $45.37$ \\
Only Discr. & $84.41$ & $41.51$ & $81.04$ & $45.52$ \\
Both (Ours) & $\textbf{85.08}$ & $\textbf{41.69}$ & $\textbf{81.41}$ & $\textbf{46.19}$ \\
\bottomrule
\end{tabularx}
\captionof{table}{Clean and robust accuracy trained with and without our method's components, on CIFAR-10. SimCLR refers to \textbf{without} SAM and Discriminator, and Ours to the method with full components.}
\label{full-table}
\end{minipage}

\textbf{Attack Settings:} We employ Projected Gradient Descent \citep{madry2017towards} as the standard adversarial attack to generate adversarial examples during training and evaluate the robustness of the defense methods in this paper. All attacks are conducted under $\ell_\infty$-norm with budget $8/255$. We use PGD-7 with step size $2/255$ for contrastive learning; PGD-10 with step size $2/255$ for robust linear evaluation; and PGD-20 with step size $0.8/255$ for final classification evaluation. We also use $\ell_2$-PGD and AutoAttack \citep{croce2020reliable} as unseen attacks.



\textbf{Baselines:} We directly compare our method with two seminal robust CL works: RoCL \citep{NEURIPS2020_robust1} and CLAE \citep{NEURIPS2020_robust2}. 
We also verify our method's effectiveness by incorporating it to ACL-DS \citep{ACL2020}, and recent methods AdvCL \citep{fan2021does}, ACL-Dyn \citep{luo2023rethinking} and ACL-AIR \citep{xu2023enhancing}.

\vspace{-2mm}
\subsection{Experimental Results}
\subsubsection{Main results}



Table \ref{main-table} shows the clean and robust accuracy under different attacks of our method and baselines, under two types of linear evaluation, LE and AT-LE, on CIFAR-10 and CIFAR-100. By encouraging sharpness-aware minimization on the benign loss, i.e., InfoNCE loss on benign examples, and taking into account the global distance between these examples and their adversarial counterparts, our method can achieve higher robust accuracy and significantly improve clean accuracy simultaneously. The successful incorporation with AdvCL indicates that our method could be used with other advanced robust CL works. To further demonstrate this, Table \ref{new-baseliens-table} shows the consistent improvement in both clean and robust accuracy of other baselines.

\subsubsection{Ablation Studies}

We then verify GSA-RSL's transferability by linearly finetuning a model pre-trained on CIFAR-10 to fit to STL-10. As shown in Table \ref{stl-table}, our method achieves a higher accuracy in all settings comparing to previous works. This observation persists when we apply our method on top of an orthogonal approach, AdvCL, further demonstrating its effectiveness.


We design the experiments according to Remark \ref{full_remark} to verify our theoretical findings. First, we show that minimizing the InfoNCE loss on the benign samples in addition to the adversarial training loss can improve both clean \textit{and} robust accuracy. These results can generally be further improved by applying sharpness-aware minimization to the optimization process. Second, adding the discriminator to the original adversarial training process leads to higher robust accuracy, albeit potentially at the expense of a small decrease in clean accuracy. Third, by combining these improvements, we arrive at our full method, which yields an even higher accuracy on most of the settings.

\textbf{Impact of the Benign Loss Term with SAM:}
We experiment with incorporating a benign loss to the traditional adversarial loss. Surprisingly, as the benign loss weight increases, we improve both the clean and robust accuracy of our model in the linear evaluation (LE) phase, with and without adversarial training (AT), cf. green and red dashed lines in Fig. \ref{fig:ablation}.
Additionally, applying SAM to the benign loss can increase the performance even further (cf. Fig. \ref{fig:SAM}). 


\textbf{Impact of the Discriminator in Training:}
By adding our discriminator to the original adversarial training scheme, we get better adversarial accuracy while trading off some clean accuracy (cf. Fig. \ref{fig:D}). This is consistent with the adversarial literature \citep{raghunathan2020understanding}. 

\textbf{Putting Everything Together:}
Table \cref{full-table} reports the effectiveness of each component and the overall improvement when all are combined. Note that the baseline with out SAM and discriminator consists of only the benign and adversarial losses. 
Both clean and robust accuracy are higher than those of this baseline.

       

       

\section{Conclusion}
Contrastive learning is a powerful technique that enables learning of meaningful features without relying on labeled information. Its initial phase focuses on learning these features, which are subsequently classified using a linear classifier trained on labeled data. Although previous theoretical studies have explored the relationship between the supervised loss in the second phase and the unsupervised loss in the first phase,
none has examined the connection between the first phase's loss and the second phase's \textit{robust} supervised loss.
To address this research gap, our paper develops rigorous theories to identify the specific components within the unsupervised loss that contribute to the robustness of the supervised loss. Lastly, we conduct 
extensive experiments to validate our theoretical findings.

\bibliographystyle{abbrv}
\bibliography{ref}

\end{document}